\documentclass{article}
\usepackage[final]{arxiv_version}

\usepackage{makeidx}

\usepackage{graphicx}

\usepackage[english]{babel}
\usepackage[utf8]{inputenc} 
\usepackage[T1]{fontenc}    
\usepackage{hyperref}       
\usepackage{url}            
\usepackage{booktabs}       
\usepackage{amsfonts}       
\usepackage{nicefrac}       
\usepackage{microtype}      
\usepackage{amsmath} 		
\usepackage {tikz}			

\title{Multi-view Generative Adversarial Networks}

\author{Mickaël Chen \\
  Sorbonne Universités, UPMC Univ Paris 06, UMR 7606, LIP6, F-75005, Paris, France\\
  \texttt{mickael.chen@lip6.fr}\\
   \And
   Ludovic Denoyer \\
  Sorbonne Universités, UPMC Univ Paris 06, UMR 7606, LIP6, F-75005, Paris, France\\
   \texttt{ludovic.denoyer@lip6.fr} \\
}

\begin{document}

\maketitle

\begin{abstract}
Learning over multi-view data is a challenging problem with strong practical applications. Most related studies focus on the classification point of view and assume that all the views are available at any time. We consider an extension of this framework in two directions. First, based on the BiGAN model, the Multi-view BiGAN (MV-BiGAN) is able to perform density estimation from multi-view inputs. Second, it can deal with missing views and is able to update its prediction when additional views are provided. We illustrate these properties on a set of experiments over different datasets.
\end{abstract}

\section{Introduction}

Many concrete applications involve multiple sources of information generating different views on the same object \citep{Cesa-Bianchi2010}. If we consider human activities for example, GPS values from a mobile phone, navigation traces over the Internet, or even photos published on social networks are different views on a particular user. In multimedia applications, views can correspond to different modalities  \citep{Atrey} such as sounds, images, videos, sequences of previous frames, etc... 

The problem of multi-view machine learning has been extensively studied during the last decade, mainly from the classification point of view. In that case, one wants to predict an output $y$ based on multiple views acquired on an unknown object $x$. Different strategies have been explored but a general common idea is based on the (early or late) fusion of the different views at a particular level of a deep architecture \citep{wang2015deep,ngiam2011multimodal,srivastava2012multimodal}.

The existing literature mainly explores problems where outputs are chosen in a discrete set (e.g categorization), and where all the views are available. An extension of this problem is to consider the density estimation problem where one wants to estimate the conditional probabilities of the outputs given the available views. As noted by \citep{mathieu2015deep}, minimizing classical prediction losses (e.g Mean square error) will not capture the different output distribution modalities. 

In this article, we propose a new model able to estimate a distribution over the possible outputs given any subset of views on a particular input. This model is based on the (Bidirectional) \textit{Generative Adversarial Networks} (BiGAN) formalism. More precisely, we bring two main contributions: first, we propose the CV-BiGAN (\textit{Conditional Views BiGAN} -- Section \ref{sec:cvgan}) architecture that allows one to model a conditional distribution $P(y|.)$ in an original way. Second, on top of this architecture, we build the Multi-view BiGANs (MV-BiGAN -- Section \ref{sec:mvgan}) which is able to both \textbf{predict when only one or few views are available}, and to \textbf{update its prediction if new views are added}. We evaluate this model on different multi-views problems and different datasets (Section \ref{sec:exp}). The related work is provided in Section \ref{sec:related} and we propose some future research directions in Section \ref{sec:conclusion}.

\section{Background and General Idea}

\subsection{Notations and Task}

Let us denote $\mathcal{X}$ the space of objects on which different views will be acquired. Each possible input $x \in \mathcal{X}$ is associated to a target prediction $y \in \mathbb{R}^n$. A classical machine learning problem is to estimate $P(y|x)$ based on the training set. But we consider instead a multi-view problem in which different views on $x$ are available, $x$ being unknown. Let us denote $V$ the number of possible views and $\tilde{x}_k$ the $k$-th view over $x$. The description space for view $k$ is $\mathbb{R}^{n_k}$ where $n_k$ is the number of features in view $k$. 
Moreover, we consider that some of the $V$ views can be missing. The subset of available views for input $x^i$ will be represented by an index vector $s^i \in \mathcal{S}=\{0,1\}^V$ so that $s^i_k=1$ if view $k$ is available and $s^i_k=0$ elsewhere. Note that all the $V$ views will not be available for each input $x$, and the prediction model must be able to predict an output given any subset of views $s \in \{0;1\}^V$. 

In this configuration, our objective is to estimate the distributions $p(y|v(s,x))$ where $v(s,x)$ is the set of views $\tilde{x}_k$ so that $s_k=1$. This distribution $p$ will be estimated based on a training set $\mathcal{D}$ of $N$ training examples. Each example is composed of a subset of views $s^i,v(s^i,x^i)$ associated to an output $y^i$, so that $\mathcal{D}=\{ \left(y^1,s^1,v(s^1,x^1) \right),...,\left(y^N,s^N,v(s^N,x^N) \right) \}$ where $s^i$ is the index vector of views available for $x^i$. Note that $x^i$ is not directly known in the training set but only observed through its associated views.

\subsection{Bidirectional Generative Adversarial Nets (BiGAN)}
We quickly remind the principle of BiGANs since our model is an extension of this technique. Generative Adversarial Networks (GAN) have been introduced by \citep{goodfellow2014generative} and have demonstrated their ability to model complex distributions. They have been used to produce compelling natural images from a simple latent distribution \citep{radford2015unsupervised,denton2015deep}. Exploring the latent space has uncovered interesting, meaningful patterns in the resulting outputs. However, GANs lack the ability to retrieve a latent representation given an output, missing out an opportunity to exploit the learned manifold. Bidirectional Generative Adversarial Networks (BiGANs) have been proposed by \citep{donahue2016adversarial} and \citep{dumoulin2016adversarially}, independently, to fill that gap. BiGANs simultaneously learn both an encoder function $E$ that models the encoding process $P_E(z|y)$ from the space $\mathbb{R}^n$ to a latent space $\mathbb{R}^Z$, and a generator function $G$ that models the mapping distribution $P_G(y|z)$ of any latent point $z \in \mathbb{R}^Z$ to a possible object $y \in \mathbb{R}^n$. From both the encoder distribution and the generator distribution, we can model two joint distributions, respectively denoted $P_E(y,z)$ and $P_G(y,z)$:
\begin{equation}
	\begin{aligned}	
		P_G(y,z) &= P(z)P_G(y|z)\\
		P_E(y,z) &= P(y)P_E(z|y)
    \end{aligned}
\end{equation}
assuming that $P(z)=\mathcal{N}(0,1)$ and $P(y)$ can be estimated over the training set by a uniform sampling. The BiGAN framework also introduces a discriminator network $D_1$ whose task is to determine whether a pair $(y,z)$ is sampled from $p_G(y,z)$ or from $p_E(y,z)$, while $E$ and $G$ are trained to fool $D_1$, resulting in the following learning problem:

\begin{equation}
	\begin{aligned}
		\min\limits_{G,E} \max\limits_{D_1} \, & \mathbb{E}_{y\sim P(y), z\sim P_E(z|y)} \left[ \log D_1(y,z) \right] \\
		+ & \, \mathbb{E}_{z\sim P(z), y\sim P_G(y|z)} \left[ 1 - \log D_1(y,z) \right]
	\end{aligned}
\label{eq:bigan}
\end{equation}

It can be shown, by following the same steps as in \citep{goodfellow2014generative}, that the optimization problem described in Equation \ref{eq:bigan} minimizes the Jensen-Shanon divergence between $P_E(y,z)$ and $P_G(y,z)$, allowing the model to learn both a decoder and a generator over a training set that will model the joint distribution of $(y,z)$ pairs. As proposed by \citep{dumoulin2016adversarially}, we consider in the following that $P_G(y|z)$ is modeled by a deterministic non-linear model $G$ so that $G(z)=y$, and $P_E$ as a diagonal Gaussian distribution $E(z)=(\mu(y), \sigma(y) )$. $G$, $\mu$ and $\sigma$ are estimated by using gradient-based descent techniques.

\subsection{General Idea}

We propose a model based on the Generative Adversarial Networks paradigm adapted to the multi-view prediction problem. Our model is based on two different principles:

\textbf{Conditional Views BiGANs (CV-BiGAN): } First, since one wants to model an output distribution based on observations, our first contribution is to propose an adaptation of BiGANs to model conditional probabilities, resulting in a model able to learn $P(y|\tilde{x})$ where $\tilde{x}$ can be either a single view or an aggregation of multiple views. If conditional GANs have already been proposed in the literature (see Section \ref{sec:related}) they are not adapted to our problem which require explicit mappings between input space to latent space, and from latent space to output space.  

\textbf{Multi-View BiGANs (MV-BiGAN): } On top of the CV-BiGAN model, we build a multi-view model able to estimate the distribution of possible outputs based on any subset of views $v(s,x)$. If a natural way to extend the Conditional BiGANS for handling multi-view is to define a mapping function which map the set of views to a representation space (see Section \ref{subsec:agg}) the resulting model has shown undesirable behaviors (see Section \ref{subsec:exp1}). Therefore, we propose to constrain the model based on the idea that adding one more view to any subset of views must decrease the uncertainty on the output distribution i.e the more views are provided, the less variance the output distribution has. This behavior is encouraged by using a Kullback-Leibler divergence (KL) regularization (see Section \ref{subsec:KL}).

\section{The Conditional BiGAN Model (CV-BiGAN)}
\label{sec:cvgan}

Our first objective is to extend the BiGAN formalism to handle an input space (e.g a single observed view) in addition to the output space $\mathbb{R}^n$. We will denote $\tilde{x}$ the observation and $y$ the output to predict. In other words, we wish to capture the conditional probability $P(y|\tilde{x})$ from a given training dataset. Assuming one possesses a bidirectional mapping between the input space and an associated representation space, ie. $P_E(z|y)$ and $P_G(y|z)$, one can equivalently capture $P(z|\tilde{x})$. The CV-BiGAN model keeps the encoder $E$ and generator $G$ defined previously but also includes an additional encoder function denoted $H$ which goal is to map a value $\tilde{x}$ to the latent space $\mathbb{R}^Z$. Applying $H$ on any value of $\tilde{x}$ results in a distribution $P_H(z|\tilde{x}) = \mathcal{N}(\mu_H(\tilde{x}),\sigma_H(\tilde{x}))$ so that a value of $z$ can be sampled from this distribution. This would then allow one to recover a distribution $P(y|\tilde{x})$. 

\begin{figure}[t]
\def\svgwidth{\linewidth}
\begin{center}
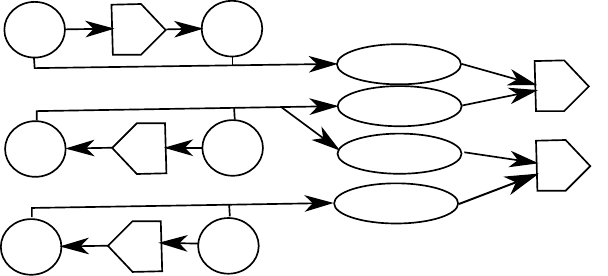
\end{center}
\caption{The CV-BiGAN Architecture. The two top levels correspond to the BiGAN model, while the third level is added to model the distribution over the latent space given the input of the CV-BiGAN. The discriminator $D_2$ is used to constraint $P(z|y)$ and $P(z|\tilde{x})$ to be as close as possible.}
\label{fig:cvgan}
\end{figure}

Given a pair $(\tilde{x},y)$, we wish a latent representation $z$ sampled from $P_H(z|\tilde{x})$ to be similar to one from $P_E(z|y)$. As our goal here is to learn $P(z|\tilde{x})$, we define two joint distributions between $\tilde{x}$ and $z$:
\begin{equation}
\begin{aligned}
P_H(\tilde{x},z)&=P_H(z|\tilde{x})P(\tilde{x})\\
P_E(\tilde{x},z)&=\sum\limits_yP_E(z|y)P(\tilde{x},y)
\end{aligned}
\end{equation}

Minimizing the Jensen-Shanon divergence between these two distributions is equivalent to solving the following adversarial problem:
\begin{equation}
\begin{aligned}
\min\limits_{E,H} \max\limits_{D_2} \, &
\mathbb{E}_{\tilde{x},y\sim p(\tilde{x},y), z\sim p_E(z|y)} \left[ \log D_2(\tilde{x},z) \right] \\
+ \, & \mathbb{E}_{\tilde{x},y\sim p(\tilde{x},y), z\sim p_H(z|x)} \left[ 1 - \log D_2(\tilde{x},z) \right]
\end{aligned}
\label{eq:cvgan}
\end{equation}

Note that when applying stochastic gradient-based descent techniques over this objective function, the probability $P(\tilde{x},y)$ is approximated by sampling uniformly from the training set. We can sample from $P_H(\tilde{x},z)$ and $P_E(\tilde{x},z)$ by forwarding the pair $(\tilde{x},y)$ into the corresponding network.

By merging the two objective functions defined in Equation \ref{eq:bigan} and \ref{eq:cvgan}, the final learning problem for our Conditionnal BiGANs is defined as: 
\begin{equation}
\begin{aligned}
\min\limits_{G,E,H} \max\limits_{D_1, D_2} \ 
&\mathbb{E}_{\tilde{x},y\sim P(\tilde{x},y), z\sim P_E(z|y)} \left[ \log D_1(y,z) \right] \\ 
+ & \, \mathbb{E}_{z\sim P(z), y\sim P_G(y|z)} \left[ 1 - \log D_1(y,z) \right] \\
+ & \, \mathbb{E}_{\tilde{x},y\sim P(\tilde{x},y), z\sim p_E(z|y)} \left[ \log D_2(\tilde{x},z) \right] \\
+ & \, \mathbb{E}_{\tilde{x},y\sim P(\tilde{x},y), z\sim P_H(z|\tilde{x})} \left[ 1 - \log D_2(\tilde{x},z) \right]
\end{aligned}
\label{eq:cvbigan}
\end{equation}

The general idea of CV-BiGAN is illustrated in Figure \ref{fig:cvgan}.

\section{Multi-View BiGAN}
\label{sec:mvgan}

\subsection{Aggregating Multi-views for CV-BiGAN}
\label{subsec:agg}
We now consider the problem of computing an output distribution conditioned by multiple different views. In that case, we can use the CV-BiGAN Model (or other conditional approaches) conjointly with a model able to aggregate the different views where $A$ is the size of the aggregation space. Instead of considering the input $\tilde{x}$, we define an aggregation model $\Psi$. $\Psi(v(s,x))$ will be the representation of the aggregation of all the available views $\tilde{x}_k$\footnote{ Note that other aggregation scheme can be used like recurrent neural networks for example.}:
\begin{equation}
\Psi(v(s,x))= \sum\limits_{k=1}^V s_k \phi_k(\tilde{x}^k)
\end{equation}
where $\phi_k$ is a function that will be learned that maps a particular view in $\mathbb{R}^{n_k}$ to the aggregation space $\mathbb{R}^A$. By replacing $\tilde{x}$ in Equation \ref{eq:cvbigan}, one can then simultaneously learn the functions $\phi_k$ and the distributions $P_H$, $P_E$ and $P_D$, resulting in a multi-view model able to deal with any subset of views.  

\begin{figure}[t]
\def\svgwidth{\linewidth}
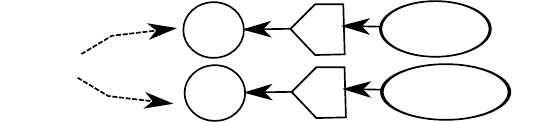
\caption{The MV-BiGAN additional components. In this example, we consider a case where only $\tilde{x}^1$ is available (top level) and a second case where both $\tilde{x}^1$ and $\tilde{x}^3$ are available. The distribution $P(z|\tilde{x}^1,\tilde{x}^3)$ is encouraged to be ''included'' in $P(z|\tilde{x}^1)$ by the KL constraint. The aggregation of the views is made by the $\phi_k$ functions that are learned conjointly with the rest of the model.}
\label{fig}
\end{figure}

\subsection{Uncertainty reduction assumption}
\label{subsec:KL}

However, the previous idea suffers  from a very high instability when learning, as it is usually noted with complex GANs architectures (see Section \ref{sec:exp}). In order to stabilize our model, we propose to add a regularization based on the idea that adding new views to an existing subset of views should reduce the uncertainty over the outputs. Indeed, under the assumption that views are consistent one another, adding a new view should allow to refine the predictions and reduce the variance of the distribution of the outputs.

Let us consider an object $x$ and two index vectors $s$ and $s'$ such that  $v(x,s) \subset v(x,s')$ ie. $\forall k, s'_k \geq s_k$. Then, intuitively, $P(x|v(x,s'))$ should be ''included'' in  $P(x|v(x,s))$. In the CV-GAN model, since $P(y|z)$ is deterministic, this can be enforced at a latent level by minimizing KL($P(z|v(x,s')$ || $P(z|v(x,s)$). By assuming those two distributions are diagonal gaussian distributions (ie. $P(z|v(x,s') = \mathcal{N}(\mu_1,\Sigma_1)$ and $P(z|v(x,s) = \mathcal{N}(\mu_2,\Sigma_2)$ where $\Sigma_k$ are diagonal matrices with diagonal elements $\sigma_{k(i)}$), the KL divergence can be computed as in Equation \ref{eq:KL} and differentiated.

\begin{equation}
\label{eq:KL}
\begin{aligned}
KL(P(z|v(x,s'))||P(z|v(x,s)))= \\ \frac{1}{2} \sum \limits_{i=1}^Z \left( -1 - \log\left(\frac{\sigma_{1(i)}^2}{\sigma_{2(i)}^2}\right) + \frac{\sigma_{1(i)}^2}{\sigma_{2(i)}^2} + \frac{(\mu_{1(i)} - \mu_{2(i)})^2}{\sigma_{2(i)}^2}  \right)
\end{aligned}
\end{equation}

Note that this divergence is written on the estimation made by the function $H$ and will act as a regularization over the latent conditional distribution.

The final objective function of the MV-BiGAN can be written as: 
\begin{equation}
\begin{aligned}
\min\limits_{G,E,H} & \max\limits_{D_1, D_2} \, \mathbb{E}_{s,x,y\sim P(s,x,y), z\sim P_E(z|y)} \left[ \log D_1(y,z) \right] \\ 
+ & \, \mathbb{E}_{z\sim P(z), y\sim P_G(y|z)} \left[ 1 - \log D_1(y,z) \right] \\
+ & \, \mathbb{E}_{s,x,y\sim P(s,x,y), z\sim P_E(z|y)} \left[ \log D_2(v(x,s),z) \right] \\
+ & \mathbb{E}_{s,x,y\sim P(s,x,y), z\sim P_H(z|v(x,s))} \left[ 1 - \log D_2(v(x,s),z) \right] \\
+ & \lambda \mathbb{E}_{x\sim P(x)} \sum\limits_{\substack{s,s' \in \mathcal{S}_x \\ \forall k, s'_k \geq s_k}} KL(H(v(x,s'))||H(v(x,s)))
\end{aligned}
\end{equation}
where $\lambda$ controls the strength of the regularization. Note that aggregation models $\Psi$ are included into $H$ and $D_2$ and can be optimized conjointly in this objective function.

\subsection{Learning the MV-BiGAN}

The different functions $E$, $G$, $H$, $D_1$ and $D_2$ are implemented as parametric neural networks and trained by mini-batch stochastic gradient descent (see Section 5.4 for more details concerning the architectures).We first update the discriminators networks $D_1$ and $D_2$, then we update the generator and encoders $G$, $E$ and $H$ with gradient steps in the opposite direction.

As with most other implementation of GAN-based models, we find that using an alternative objective proposed by \citep{goodfellow2014generative} for $E$, $G$ and $H$ instead leads to more stable training. The new objective consist of swapping the labels for the discriminators instead of reversing the gradient. We also find that we can update all the modules in one pass instead of taking alternate gradient steps while obtaining similar results.

Note that the MV-BiGAN model is trained based on datasets where all the $V$ views are available for each data point. In order to generate examples where only subsets of views are available, the ideal procedure would be to consider all the possible subsets of views. Due to the number of data points that would be generated by such a procedure, we build random sequences of incremental sets of views and enforce the KL regularization over successive sets.

\section{Experiments}
\label{sec:exp}

We evaluate our model on three different types of experiments, and on two differents datasets. The first dataset we experiment on is the MNIST dataset of handwritten digits. The second dataset is the CelebA \citep{liu2015faceattributes} dataset composed of both images of faces and corresponding attributes. The MNIST dataset is used to illustrate the ability of the MV-BiGAN to handle different subset of views, and to update its prediction when integrating new incoming views. The CelebA dataset is used to demonstrate the ability of MV-BiGAN to deal with different types (heterogeneous) of views.

\subsection{MNIST, 4 views}
\label{subsec:exp1}

\begin{figure}[ht]
\begin{center}
\includegraphics[width=0.8\linewidth]{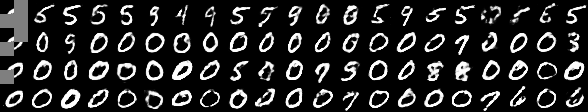}\\
\includegraphics[width=0.8\linewidth]{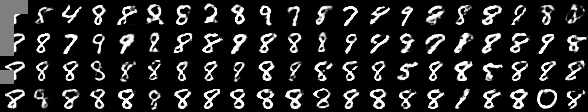}\\
\includegraphics[width=0.8\linewidth]{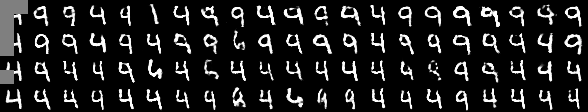}\\
\includegraphics[width=0.8\linewidth]{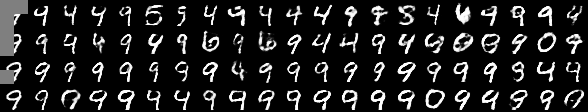}  \\
\includegraphics[width=0.8\linewidth]{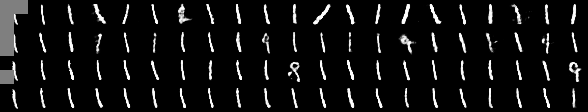}  \\
\includegraphics[width=0.8\linewidth]{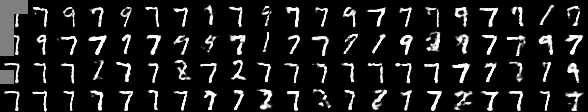}
\end{center}
\caption{Results of the MV-BiGAN on sequences of 4 different views. The first column corresponds to the provided views, while the other columns correspond to outputs sampled by the MV-BiGAN.}
\label{fig1}
\end{figure}

\begin{figure}[ht]
\begin{center}
\includegraphics[width=0.8\linewidth]{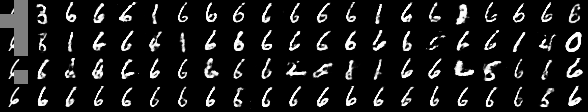}\\
\includegraphics[width=0.8\linewidth]{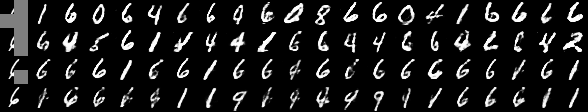}
\end{center}
\caption{Comparaison between MV-BiGAN with (top) and without (bottom) KL-constraint.}
\label{fig2}
\end{figure}

We consider the problem where 4 different views can be available, each view corresponding to a particular quarter of the final image to predict -- each view is a vector of $\mathbb{R}^{(14 \times 14)}$. The MV-BiGAN is used here to recover the original image. The model is trained on the MNIST training digits, and results are provided on the MNIST testing dataset. 

Figure \ref{fig1} illustrates the results obtained for some digits. In this figure, the first column displays the input (the subset of views), while the other columns shows predicted outputs sampled by the MV-BiGAN. An additional view is added between each row. This experiment shows that when new views are added, the diversity in the predicted outputs decreases due to the KL-contraint introduced in the model, which is the desired behavior i.e more information implied less variance. When removing the KL constraint (Figure \ref{fig2}), the diversity still remains important, even if many views are provided to the model. This show the importance of the KL regularization term in the MV-BiGAN objective.

\subsection{MNIST, sequence of incoming views}

\begin{figure}[ht!]
\begin{center}
\includegraphics[width=0.8\linewidth]{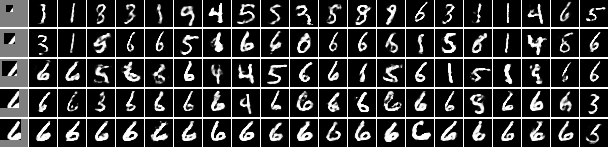}\vspace{0.1cm}
\includegraphics[width=0.8\linewidth]{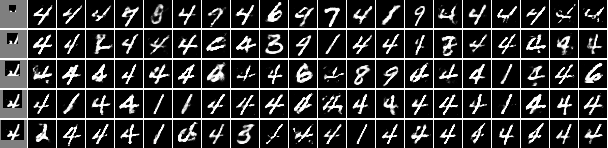}\vspace{0.1cm}
\includegraphics[width=0.8\linewidth]{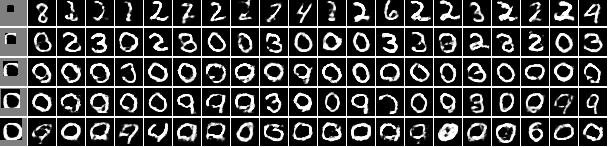}
\end{center}
\caption{MV-BiGAN with sequences of incoming views. Here, each view is a $28 \times 28$ matrix (values are between $0$ and $1$ with missing values replaced by $0.5$). }
\label{fig3}
\end{figure}

We made another set of experiments where the views correspond to images with missing values (missing values are replaced by $0.5$). This can be viewed as a data imputation problem -- Figure \ref{fig3}. Here also, the behavior of the MV-BiGAN exhibits interesting properties: the model is able to predict the desired output as long as enough information has been provided. When only non-informative views are provided, the model produces digits with a high diversity, the diversity decreasing when new information is added.

\subsection{CelebA, integrating heterogeneous information}

\begin{figure}
\centering
\begin{tabular}{cccc}
\includegraphics[width=0.05\linewidth]{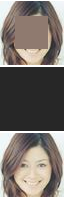} & 
\includegraphics[width=0.35\linewidth]{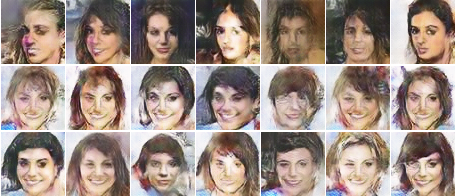} & 
\includegraphics[width=0.05\linewidth]{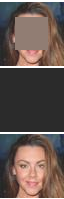} & 
\includegraphics[width=0.35\linewidth]{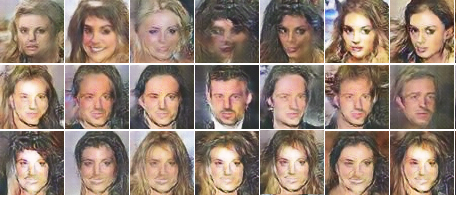} 
\end{tabular}
\caption{Results obtained on the CelebA dataset for two examples. The first line corresponds to the images generated based on the attribute vector, the second line corresponds to images generated based on the incomplete face, the third line corresponds to the images generated based on the two views. The groundthruth face is given in the bottom-left corner, while the incomplete face is given in the top-left corner.} 
\label{celeb}
\end{figure}

At last, the third experiment aims at measuring the ability of MV-BiGAN to handle heterogeneous inputs. We consider two views: (i) the attribute vector containing information about the person in the picture (hair color, sex, ...), and (ii) a incomplete face. Figure \ref{celeb} illustrates the results obtained on two faces. The first line corresponds to the faces generated based on the attribute vector. One can see that the attribute information has been captured by the model: for example, the sex of the generated face is constant (only women) showing that MV-BiGan has captured this information from the attribute vector. The second line corresponds to the faces generated when using the incomplete face as an input. One can also see that the generated outputs are ''compatible'' with the incomplete information provided to the model. But the attribute are not considered (for example, women and men are generated). At last, the third line corresponds to images generated based on the two partial views (attributes and incomplete face) which are close to the ground-truth image (bottom left). Note that, in this set of experiments, the convergence of the MV-BiGAN was quite difficult to obtain, and the quality of the generated faces is still not satisfying. 

\subsection{Implementation details}

All models are optimized using Adam with standard hyperparameters $\beta_1 = 0.5$, $\beta_2 = 10^{-3}$ and a learning rate of $2\cdot 10^{-5}$.
All hidden layers in generator or encoder networks are followed by a rectified linear unit. In discriminator networks, leaky rectified linear units of slope $0.2$ are used instead.
Latent representations $(\mu, \log (\sigma ^2))$  are of size 2$\times$128.

For \textbf{MNIST experiments}, the generator function $G$ has three hidden fully connected layers. The second and the third hidden layers are followed by batch normalizations. The output layer uses a sigmoid.

The aggregation model $\Psi$ is a sum of mapping functions $\phi_k$. Each $\phi_k$ is a simple linear transformation. 
The encoding functions E and H are both neural networks that include an aggregation network followed by two fully connected layers. A batch normalization is added after the second layer. They output a pair of vectors $(\mu, \log (\sigma ^2))$. The output layers has a tanh for $\mu$ and a negative exponential linear unit for $\log \sigma ^2$.

The discriminator $D_1$ has three fully connected layers with batch normalization at the third layer. A sigmoid is applied to the outputs. The vector $z$ is concatenated to the representation at the second layer.

The discriminator $D_2$ is similar to E and H except it uses  a sigmoid at the output level. $z$ is concatenated directly to the aggregation vector $\Psi(v(x,s))$.

All hidden layers and the aggregation space are of size 1500. $\lambda$ is set to $1\cdot10^{-5}$. Minibatch size is set to 128. The models have been trained for 300 epochs.

For \textbf{CelebA experiments}, the generator function $G$ is a network of transposed convolution layers described in table \ref{table:conv}.

\begin{table}[h]
\centering
\begin{tabular}{@{}rllllll@{}} \toprule
Operation              & Kernel       & Strides      & Padding & Feature maps & BN      & Nonlinearity \\ \midrule
Convolution            & $4 \times 4$ & $2 \times 2$ & $1 \times 1$ & $64$         & $\times$ & Leaky ReLU \\
Convolution            & $4 \times 4$ & $2 \times 2$ & $1 \times 1$ & $128$        & $\surd$  & Leaky ReLU \\
Convolution            & $4 \times 4$ & $2 \times 2$ & $1 \times 1$ & $256$        & $\surd$  & Leaky ReLU \\
Convolution            & $4 \times 4$ & $2 \times 2$ & $1 \times 1$ & $512$        & $\surd$  & Leaky ReLU \\
Convolution            & $4 \times 4$ & $1 \times 1$ & 				& output size  & $\times$ & Linear     \\
\midrule
Transposed convolution            & $4 \times 4$ & $1 \times 1$ & 			 & $512$        & $\surd$ & ReLU \\
Transposed convolution            & $4 \times 4$ & $2 \times 2$ & $1 \times 1$ & $256$         & $\surd$ & ReLU \\
Transposed convolution            & $4 \times 4$ & $2 \times 2$ & $1 \times 1$ & $128$        & $\surd$  & ReLU \\
Transposed convolution            & $4 \times 4$ & $2 \times 2$ & $1 \times 1$ & $64$        & $\surd$  & ReLU \\
Transposed convolution            & $4 \times 4$ & $2 \times 2$ & $1 \times 1$ & $3$        & $\times$  & Tanh \\
\bottomrule
\end{tabular}
\label{table:conv}
\caption{Convolution architectures used in our experiments on the CelebA dataset. The top part is used for encoding images into the aggregation space. The bottom part is used in $G$ to generate images from a vector $z$.}
\end{table}

The mapping functions $\phi_k$ for images are convolution networks (Table \ref{table:conv}). For attribute vectors, they are linear transformations. E and H are neural networks with one hidden layer on top of the aggregation model. The hidden layer is followed by a batch normalization. The output layer is the same as in the MNIST experiments.
The discriminator $D_1$  is a transposed convolution network followed by a hidden fully connected layer before the output layer. $z$ is concatenated at the hidden fully connected level.
As in the MNIST experiments, the discriminator $D_2$ is similar to E and H, and $z$ is concatenated directly to the aggregation vector $\Psi(v(x,s))$.
Aggregation space is of size 1000. $\lambda$ is set to $1\cdot10^{-3}$, and mini-batch size is 16. The model has been trained for 15 epochs.

\section{Related work}
\label{sec:related}
\textbf{Multi-view and Representation Learning: } Many application fields naturally deal with multi-view data with true advantages. For example, in the multimedia domain, 
dealing with a bunch of views is usual \citep{Atrey}: text, audio, images (different framings from videos) are starting points of these views. Besides, multimedia learning tasks from multi-views led to a large amount of fusion-based ad-hoc approaches and experimental results. The success of multi-view supervised learning approaches in the multimedia community seems to rely on the ability of the systems to deal with the complementary of the information carried by each modality. Comparable studies are of importance in many domains, such as bioinformatics \citep{Sokolov2011multiview}, 
speech recognition \citep{Arora2012kernel,Koco2012applying}, signal-based multimodal integration \citep{Wu1999multimodal}, gesture recognition \citep{Wu2013fusing}, etc. 

Moreover, multi-view learning has been theoretically studied mainly under the semi-supervised setting, but only with two facing views \citep{Chapelle2006semi,Sun2013survey,Sun2014pac,Johnson2015semi}.  In parallel, ensemble-based learning approaches have been theoretically studied, in the supervised setting: many interesting results should concern multi-view learning, as long as the ensemble is built upon many views \citep{Rokach2010ensemble,Zhang2011novel}. 
From the representation learning point of view, recent models are based on the incorporation of some ''fusion'' layers in the deep neural network architecture as in \citep{ngiam2011multimodal} or \citep{srivastava2012multimodal} for example. Some other interesting models include the multiview perceptron\citep{zhu2014multi}.

\textbf{Estimating Complex Distributions: } While deep learning has shown great results in many classification task for a decade, training deep generative models still remains a challenge. Deep Boltzmann Machines \citep{salakhutdinov2009deep} are un-directed graphical models organized in a succession of layers of hidden variables. In a multi-view setting, they are able to deal with missing views and have been used to capture the joint distribution in bi-modal text and image data \citep{srivastava2012multimodal,sohn2014improved}. 
Another trend started with denoising autoencoder \citep{vincent2008extracting}, which aims to reconstruct a data from a noisy input have been proved to possess some desirable properties for data generation \citep{bengio2013generalized}. The model have been generalized under the name Generative Stochastic Networks by replacing the noise function $C$ with a mapping to a latent space \citep{thibodeau2014deep}. Pulling away from the mixing problems encountered in previous approaches, Variational Autoencoders \citep{kingma2013auto} attempts to map the input distribution to a latent distribution which is easy to sample from. The model is trained by optimizing a variational bound on the likelihood, using stochastic gradient descent methods. The Kullback-Leibler regularizer on the latent Gaussian representations used in our model is reminiscent of the one introduced in the variational lower bound used by the VAE.

The BiGAN model \citep{donahue2016adversarial,dumoulin2016adversarially} that serves as a basis for our work is an extension of the Generative Adversarial Nets \citep{goodfellow2014generative}. A GAN extension that captures conditional probabilities (CGAN) has been proposed in \citep{mirza2014conditional}. However, as noted by \citep{mathieu2015deep} and \citep{pathak2016context}, they display very unstable behavior. More specifically, CGAN have been able to generate image of faces conditioned on an attribute vector \citep{gauthier2014conditional}, but fail to model image distribution conditioned on a part of the image or on previous frames.
In both CGAN and CVBiGAN, the generation process uses random noise to be able to generate a diversity of outputs from the same input. However, in a CGAN, the generator concatenate an independent random vector to the input while CV-BiGAN learns a stochastic latent representation of the input.
Also, some of the difficulties of CGAN in handling images as both inputs $\tilde{x}$ and outputs $\tilde{y}$ stem from the fact that CGAN's discriminator directly compares $\tilde{x}$ and $y$. In CV-BiGAN, neither discriminators has access to both $\tilde{x}$ and $y$ but only to a latent representation $z$ and either $\tilde{x}$ or $y$.

\section{Conclusion and Perspectives}
\label{sec:conclusion}

We have proposed the CV-BiGAN model for estimating conditional densities, and its extension MV-BiGAN to handle multi-view inputs. The MV-BiGAN model is able to both handle subsets of views, but also to update its prediction when new views are added. It is based on the idea that the uncertainty of the prediction must decrease when additional information is provided, this idea being handled through a KL constraint in the latent space. This work opens different research directions. The first one concerns the architecture of the model itself since the convergence of MV-BiGAN is still difficult to obtain and has a particularly high training cost. Another direction would be to see if this family of model could be used on data streams for anytime prediction.

\section*{Acknowledgments}
This  work  was  supported  by  the  French  project
LIVES  ANR-15-CE23-0026-03.

\bibliography{conference}
\bibliographystyle{arxiv_version}







\end{document}